\documentclass[10pt,twocolumn]{article} 
\usepackage{times}
\usepackage{graphicx}
\usepackage{amssymb}
\usepackage{url,hyperref}
\usepackage{listings}
\usepackage{multirow}
\usepackage{algpseudocode}
\usepackage{algorithm}
\usepackage{graphicx}
\usepackage{newtxmath}

\graphicspath{ {./images/} }

\title{Local Feature Matching with Transformers for low-end devices.\\ LoFTR method adaptation approach.}
\author{
	Kyrylo Kolodiazhnyi \\
	rotate@ukr.net 
}
\date{\today}

\begin{document}
\maketitle

\begin{abstract}
{\it
LoFTR \cite{sun2021loftr} is an efficient deep learning method for finding appropriate local feature matches on image pairs. This paper reports on the optimization of this method to work on devices with low computational performance and limited memory. The original LoFTR approach is based on a ResNet \cite{he2015deep} head and two modules based on Linear Transformer \cite{wang2020linformer} architecture. In the presented work, only the coarse-matching block was left, the number of parameters was significantly reduced, and the network was trained using a knowledge distillation technique. The comparison showed that this approach allows to obtain an appropriate feature detection accuracy for the student model compared to the teacher model in the coarse matching block, despite the significant reduction of model size. Also, the paper shows additional steps required to make model compatible with NVIDIA TensorRT runtime, and shows an approach to optimize training method for low-end GPUs.

Code and demo are available in the following repository: \url{https://github.com/Kolkir/Coarse_LoFTR_TRT}
}
\end{abstract}	
	
\section{Introduction}
Recently, neural network architecture based on the Transformer approach \cite{vaswani2017attention}, has taken a leading role in solving computer vision problems. And now Transformers based models are applied not only to such classical tasks as classification and detection of objects in images, but also to more specific ones, for example feature matching, depth estimation and so on. Solutions for such kind of problems are often used for intrinsic and extrinsic camera parameter estimation, in positioning or SLAM tasks. The crucial data for camera parameter estimation are feature correspondences on two images. The classical approach for the feature matching on images is the feature descriptors computation and their comparison, for this purpose methods SIFT \cite{Lowe:2004:DIF:993451.996342}, SURF \cite{Bay2008346}, ORB \cite{inproceedings}, etc. are used. 

The LoFTR approach uses self and cross-attention transformer layers to find feature point matches in images. The transformer's global receptive field allows this approach to detect sparse feature correspondences even in low-textured parts of images where descriptor-based approaches usually fail to detect feature differences.

However, most transformer-based approaches require significant computational resources for both training and inference, due to the quadratic complexity of computations and required memory in the layers. Therefore, it's difficult to adapt them for running on low-end devices with limited computational resources. Various modifications of the transformer architecture have being developed to solve the problem of high computational complexity. The LoFTR approach uses the Linear Transformer \cite{wang2020linformer} method, which proposes to reduce the computational complexity to \(O(N)\) by replacing the exponential kernel used in attention layers with an alternative kernel \(sim(Q, K) = \varphi(Q) \cdot \varphi(K)^T\) where \(\varphi(\cdot) = elu(\cdot) + 1\). This approach has shown good computational performance improvement and the memory consumption reduction for computer vision tasks. It's important since such type of tasks has the sequence length equals to the number of pixels in the input image. Using Linear Transformer allowed to perform feature matching for 640x480 images with acceptable performance on high-end GPUs. However, changes in the architecture is still not enough to use the transformer based approaches for less computational effective devices.

Therefore, there are engineering methods for adapting complex models to the requirements of low-end devices  \cite{fournier2021practical}. The main ones are Quantization, Pruning \cite{liang2021pruning}, and Knowledge Distillation \cite{hinton2015distilling}.

Quantization is the bitwidths lowering of the data type used for calculation and weights storing. Most often floating point calculations converts to 16bit float or 8bit integer types. To achieve accuracy comparable to the original model, this approach usually requires a special training process that takes into account the downsizing or an additional model calibration followed after downsizing. Unfortunately, this kind of optimization is often not available on consumer-level or embedded GPUs, and it's implementation can be found only in high-end GPUs. However, for devices based on CPUs, this method is available and can provide good results.  

Pruning is a method of removing network parameters that do not contribute much to the resulting accuracy. Often a suitable condition for removal is that the weights are close to zero. The resulting model may require much less memory and can be more efficient in inference. There are many pruning types, but two following major types can be distinguished: the structured pruning, when symmetric blocks of weights are removed, for example layers, and  the unstructured pruning, when the removed blocks may be of different shapes. Since this approach changes the model architecture,the manual tuning is often required to be done to restore the normal model work. The structured approach may be preferable, as it makes fewer changes to the global architecture, and it's easier or may be even unnecessary to restore the model operation. However, popular deep-learning frameworks  most often implement the unstructured method. Adapting network operation after applying unstructured pruping to complex models can be a non-trivial task that will require much time to solve, and since the method does not guarantee a stable result, it is not always appropriate to apply it.

Knowledge Distillation is a method of training a model with a teacher's help. Teacher can be a network with the same architecture but with bigger number of parameters or a network with another architecture. Most often training is performed with complex loss function where there is a component that takes into account errors on a target dataset and components that take into account teacher's knowledge. Knowledge elements that are transferred to the student models can be output values of certain layers in the teacher network, for example, it may be logits that precede softmax in classification. It is also possible to use internal layer output values of the teacher network \cite{distill2021}. This method shows good results for training more compact networks while maintaining the required accuracy, but there is no standard approach for organizing such a process. And success depends on the right knowledge transfer technique, the well-chosen loss function, and the student model architecture .    

As shown above, there is no single approach for optimizing deep learning models for use on low-end devices. Therefore, specialized solutions for specific architectures are usually developed. And this paper presents an optimization approach for the LoFTR feature matching method.

\section{Approach}
The main idea of presented approach is to significantly reduce the number of model parameters and the knowledge transfer from the original model. For this purpose, it was decided to leave only the one transformer block for coarse feature matching, although the original model contains a second block for the fine match. Also, the manual iterative  selection of fewer layers in all model blocks was done. The knowledge distillation loss function was selected, and a smaller training dataset was used. However, ground-truth feature points matches were also determined using depth maps. The training process was developed to use the Automatic Mixed Precision(AMP) technology and the gradient accumulation approach to save memory and speed up computation. The source code was adapted to compile in NVIDIA TensorRT \cite{tensorrt} engine format.  
The NVIDIA Jetson Nano \cite{jetson} with working memory size of 2Gb was chosen as a target device. And as a training platform the desktop based on Intel i5 proccessor and Nvidia GTX 1060 6Gb GPU was chosen.

\subsection{Code update for the target runtime}
Since the target platform has a very limited memory due to the fact that about a third of it is always taken by the operating system and more than a third can be taken by the deep learning framework runtime such as PyTroch or TensoFlow. So only a very limited amount of memory remains for a particular task. Therefore, it was decided to use NVIDIA TensorTR as a model runtime, because this technology requires significantly less memory for the models to run.
The original LoFTR model is written in Python, using PyTorch as a deep learning framework. To create TensorTR model there were two possibilities, the one was to use the Torch-TensorRT \cite{trtorch} compiler and the second was to convert the model in the ONNX \cite{onnx} format and then compile it using NVIDIA TensorTR SDK. The first option could not be applied due to limited resources of the target platform, because the compilation into the TensorTR format with Torch-TensorRT implies running it on the target device for real-time optimizations. Experimentally it was found that compiling ONNX requires less resources and is possible on the target device, so the second option was chosen.
However, the TensorRT 8.2 version available at the time of the study did not contain implementation of all the high-level functions used in the original PyTorch version of LoFTR. Therefore, the program code of the model was adapted to use the available functions. Most of the code changes were to rewrite constructions based on the einsum technique. An example of such a code replacement is shown below. The following code sample shows the original piece of code.  

\begin{lstlisting}[ basicstyle=\small, caption={Original code with EinSum notation function},captionpos=b, label={lst:einsumorigin}, language=Python]
KV = torch.einsum("nshd,nshv->nhdv", K, values)
\end{lstlisting}

And the following code snippet shows the similar algorithms, but built from more elementary operations.

\begin{center}
\begin{minipage}{\linewidth}
\begin{lstlisting}[ basicstyle=\small,caption={The code adapted for TensorRT compatibility},captionpos=b, label={lst:einsumreplace}, language=Python]
k = K.view(-1, self.dim).unsqueeze(2)
v = values.view(-1, self.dim).unsqueeze(1)
kv = torch.bmm(k, v)
kv = kv.reshape(-1, v_length, self.nheads, self.dim, self.dim)
KV = kv.sum(dim=1)
\end{lstlisting}
\end{minipage}
\end{center}

You can see from the example that converting a fairly simple source code \ref{lst:einsumorigin} into the one compatible with the  target platform \ref{lst:einsumreplace} is not a simple operation and cannot be done by automated tools.

\subsection{Model size optimization}
It was decided to select the model size first to achieve acceptable performance on the target device i.e. choose the number and size of layers in blocks. For this purpose, a demo application that searches for feature matches on the real-time webcam images was developed. The performance has been estimated by the number of FPS during rendering corresponding matches. Then with the help of this application, the model configuration shown in the table \ref{table:modelsizes} was iteratively chosen.

\begin{table}[h]
\begin{center}
\small
\begin{tabular}{|ll|l|l|}
	\hline
	\multicolumn{2}{|l|}{}                                           & Orininal & Reduced \\ \hline
	\multicolumn{2}{|l|}{resolution}                                 & 8              & 16            \\ \hline
	\multicolumn{1}{|l|}{\multirow{3}{*}{ResNet+FPN}}  & initial dim & 128            & 8             \\ \cline{2-4} 
	\multicolumn{1}{|l|}{}                             & blocks      & 3              & 4             \\ \cline{2-4} 
	\multicolumn{1}{|l|}{}                             & sizes       & 128, 196, 256  & 8, 16, 32, 32 \\ \hline
	\multicolumn{1}{|l|}{\multirow{4}{*}{Transformer}} & model D     & 256            & 32            \\ \cline{2-4} 
	\multicolumn{1}{|l|}{}                             & ffn D       & 256            & 32            \\ \cline{2-4} 
	\multicolumn{1}{|l|}{}                             & heads num   & 8              & 1             \\ \cline{2-4} 
	\multicolumn{1}{|l|}{}                             & layers      & 8              & 4             \\ \hline
\end{tabular}
\end{center}
\caption{Models parameter sizes.}
\label{table:modelsizes}
\end{table}

This configuration provides the performance shown in the table \ref{table:performance}.

\begin{table}[h]
\begin{center}
\small
\begin{tabular}{|l|l|l|}
	\hline
	& Jetson Nano 2Gb & GTX 1060 6Gb   \\ \hline
	Original FPS &                        & \textgreater 1 \\ \hline
	Reduced FPS  & $\sim$5                & $\sim$45       \\ \hline
\end{tabular}
\end{center}
\caption{Model performances.}
\label{table:performance}
\end{table}

The authors of the original model reported that the full model handles a pair of 640×480 images in 116 ms on an RTX 2080Ti, which is about 8 FPS \cite{sun2021loftr}. The table \ref{table:modelparams} shows the change in the number of parameters.

\begin{table}[h]
\begin{center}
\begin{tabular}{|l|l|l|}
	\hline
	& Parameters number & Parameters size, Mb \\ \hline
	Original & 27,950,897        & 106.624             \\ \hline
	Reduced  & 2,257,022         & 8.61                \\ \hline
\end{tabular}
\end{center}
\caption{Number and size of model parameters.}
\label{table:modelparams}
\end{table}

You can see from the tables, that the original model was significantly reduced in the size to achieve an acceptable performance on the target device.

\subsection{Training dataset}
In this work, the LoFTR model trained on outdoor scenes using the MegaDepth \cite{li2018megadepth} dataset was used for knowledge distillation. This dataset includes data from 196 locations taken from different angles of view and reconstructed by COLMAP SfM/MVS \cite{schoenberger2016mvs} \cite{schoenberger2016sfm}. And the minimum MegaDepth dataset is about 200GB. Since it is quite a big amount of data, it was decided to use a smaller set of training data, which could be effectively handled on a desktop with Intel® Core™ i5-7600K CPU @ 3.80GHz × 4 with 16 GB of memory and NVIDIA GeForce GTX 1060 6GB. The BlendedMVS \cite{yao2020blendedmvs} was chosen as a such dataset. This dataset contains  17k MVS training samples covering 113 different scenes, and the version with the minimum available size with low resolution (768 x 576) is 27.5GB. Each scene contains object images taken from various viewpoints, a depth map, internal and external camera parameters for each item. Also for each scene, there is a list of stereo pair images.
Since the selected dataset does not contain data about feature matches for image pairs, a ground-truth generation technique similar to the one used for training original LoFTR was used in this work too. To create a set of  ground-truth feature matches the following algorithm was performed:

\begin{algorithmic}[1]
	\State The depth values for corresponding 2D image coordinates with step size 16px are selected for the first image.
	\State 3D coordinates for the point in the stereo camera world space are calculated by applying the inverse projection using the depth value and corresponding camera matrix parameters.
	\State 3D points are projected into the 2D image space of the second camera using the corresponding matrices.
	\State The pair of points is selected as a ground-truth match if the depth difference with the second depth map for the projected point does not exceed a given threshold value.
\end{algorithmic}

On the one hand, this technique allows us to get a set of feature matches that don't relate to their geometric or textured characteristics, for example, as in the dataset used for training SuperPoint \cite{detone2018superpoint}, but on the other hand, due to insufficient resolution of depth maps, we can get incorrect matches due to overlapping objects on 2D projections.

\subsection{Knowledge distillation}
The method proposed in \cite{hinton2015distilling}  was chosen as a knowledge distillation approach for the current work. The reason for this choice is the fact that the transformer-based coarse matching block essentially solves the classification problem for a pair of points. And the result of its work is a probabilistic estimate of the image feature matches. To calculate the probabilistic estimate for feature matches the current LoFTR implementation uses the dual-softmax operator \cite{tyszkiewicz2020disk} \cite{rocco2018neighbourhood}. At first, for the transformer output  values, the score matrix \eqref{eqn:scorematrix} is calculated.
\begin{equation}\label{eqn:scorematrix}
S(i,j) = \frac{1}{\tau} \langle \tilde{F}^{A}_{tr}(i), \tilde{F}^{B}_{tr}(j)\rangle 
\end{equation}
Where the $\tilde{F}^{A}_{tr}$ and $\tilde{F}^{B}_{tr}$ are feature point values of both images, which were obtained from the ResNet+FPN convolutional head and then processed by the LoFTR module. 
For obtaining the probability of points match $P_{c}$\eqref{eqn:matchprop} softmax is applied in both directions.
\begin{equation}\label{eqn:matchprop}
	P_{c}(i,j) = softmax(S(i,\cdot))_{j} \cdot softmax(S(\cdot, j))_{i} 
\end{equation}
Matrices $S$\eqref{eqn:scorematrix} are used to transfer knowledge from the teacher model to the student model, using the Kullback-Leibler divergence loss with the soft targets as a component of the global loss function .

\begin{equation}\label{eqn:distillloss}
	L_{distill} =\sum P_{student}\log \left({\frac {P_{student}}{Q_{teacher}}}\right) \cdot t^2
\end{equation}

\begin{equation}\label{eqn:studentprob}
	P_{student,i} = \log \left( {\frac {e^{\frac {S_{student,i}}{t}}}{\sum _{j}e^{\frac {S_{student,j}}{t}}}} \right)
\end{equation}

\begin{equation}\label{eqn:teacherprob}
	P_{teacher,i} = \log \left(  {\frac {e^{\frac {S_{teacher,i}}{t}}}{\sum _{j}e^{\frac {S_{teacher,j}}{t}}}} \right)
\end{equation}

The parameter $t$ is called temperature, it's equal 5 in this work. It was chosen iteratively by $L_{distill}$ decrease estimation during the training process. The selected value showed an optimal increase of entropy and correspondingly a better knowledge transfer. 
The full loss function \eqref{eqn:fullloss} for the trained model is a weighted sum that consists of two parts, one is the already considered $L_{distill}$ and the other one $L_{target}$ is a cross-entropy loss calculated for network output and ground-truth feature matches.

\begin{equation}\label{eqn:fullloss}
	L = L_{distill} \cdot c_{d} + L_{target} \cdot c_{t} 
\end{equation}

Where $c_{d}$ and $c_{t}$ are weighting factors, in this paper they are equal to $c_{t} = 0.3$ and $c_{t} = 0.7$. Such values were chosen to compensate the relatively small values of $L_{distill}$ and to make the contributions of both parts more equal.

\subsection{Training process optimization}
The overall training process with knowledge distillation was optimized for the limitations of low-performance hardware. To accelerate gradients calculation and to reduce memory consumption, the Automatic Mixed Precision(AMP) technology was used since its implementation was available in the PyTorch deep learning framework. The essence of the technology is that some operations required for gradients calculation use float32 and another part use float16 data types. For example, convolutional and linear operations are faster to compute using float16. And other operations such as reduction require the use of a float32 range. This technology allows us to automatically select appropriate data types for all the operations involved in the model training. Its use allowed to significantly reduce the memory consumption for the model ResNet+FPN head. However, numerical computation problems for small gradients values were noted for AMP technology. So to stabilize the loss functions calculations amplification factors were added.

Despite the use of AMP, the maximum possible batch for training on the GTX 1060 was 4 pairs of 640x480 images. Therefore, to increase the batch size the gradient accumulation approach was applied. It means that the big batch is divided into $n$ series of smaller ones. For each series, the forward and reverse cycles are done, and resulting gradients values are not cleared, but summed up. Where $n = \frac{BigBatchSize}{SmallBatchSize}$. At each iteration, the loss function value is multiplied by the scaling factor $\frac{1}{n}$ before gradient calculation. And the network parameters are updated only after all $n$ iterations have passed, and then gradients are zeroed. So the training with a bigger batch was simulated using this technique. In this work, the virtual batch size was equal to 32. Although, in reality, the hardware processed a series of 8 batches each of size 4. The gradient accumulation technique doesn't implement the exact correspondence of the use of real large-size batches, so the loss and gradient values will be different for these two approaches.

Also, it was noticed that applying the learning rate scheduler can significantly speed up the training process. In this work, the AdamW \cite{loshchilov2019decoupled} optimization algorithm with standard parameters was used. The initial learning rate value equals $1 \cdot 10^{-3}$ and was multiplied by $1 \cdot 10^{-3}$ every 15 epochs.

The epoch size was chosen to be 5000 pairs of images, which were randomly selected from the original dataset.

\section{Experiments}
\begin{figure}[ht]
\includegraphics[width=\linewidth]{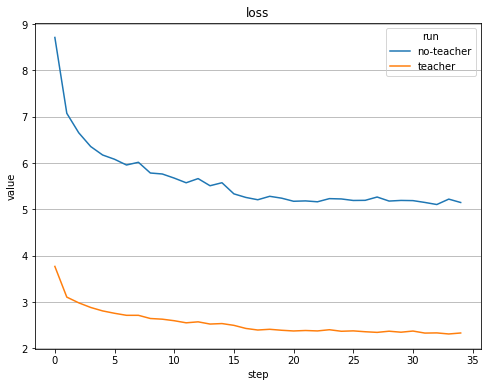}
\caption{Training loss values}
\label{img:loss}
\end{figure}

The following results were obtained as a result of applying the described approach to training a smaller network to solve the feature matching problem. The graph \ref{img:loss} shows the loss function value relation to the training duration. The graph shows the loss function values for training with and without a teacher. This graph clearly shows that the absolute loss function values are significantly smaller and the learning process itself is more stable when the training is performed with the teacher. 

\begin{figure}[ht]
	\includegraphics[width=\linewidth]{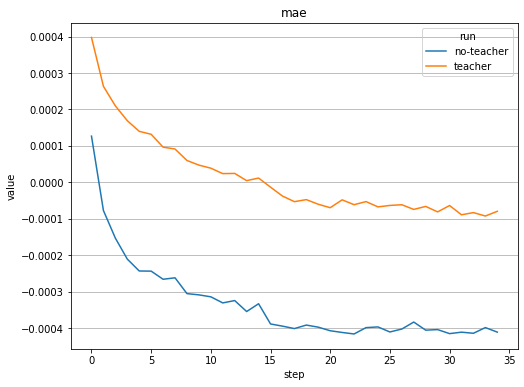}
	\caption{Training mean absolute errors}
	\label{img:mae}
\end{figure}

We can also consider the graph \ref{img:mae} that shows the dependency of Mean Absolute Error (MAE) from the training duration. It shows the average difference between predicted feature matching scores and the ground-truth values. We can see that the MAE values are much closer to zero when training is done with a teacher. We can assume that training a smaller network without a teacher makes it less confident in its results. However, at the same time, this graph \ref{img:loss} shows that the chosen model architecture is capable to learn without a teacher, but it will probably take a longer time to achieve comparable results and will require lower threshold values to determine the most significant matches. 

\begin{figure}[ht]
	\includegraphics[width=\linewidth]{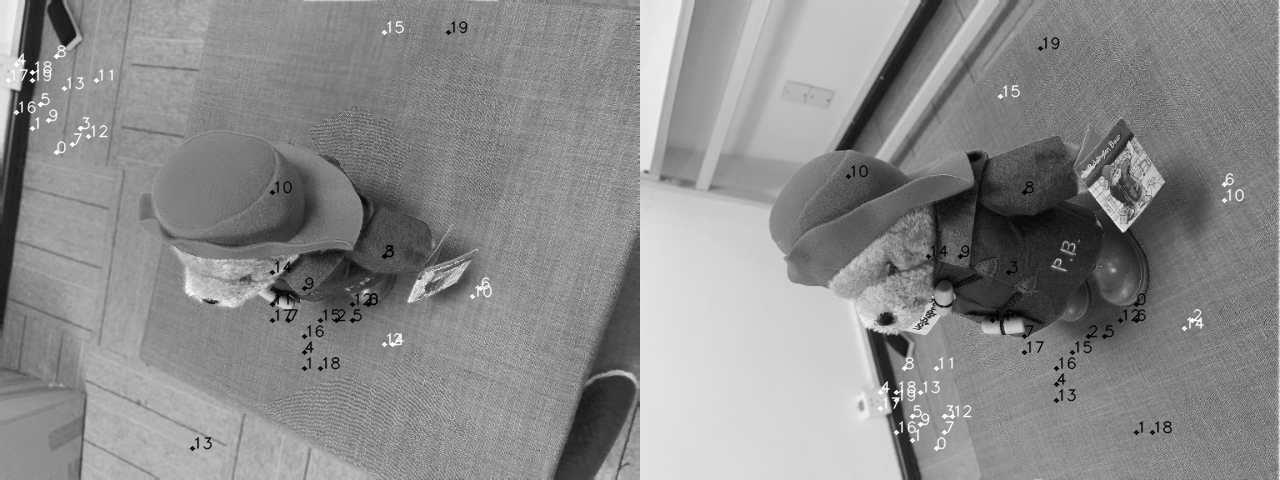}
	\includegraphics[width=\linewidth]{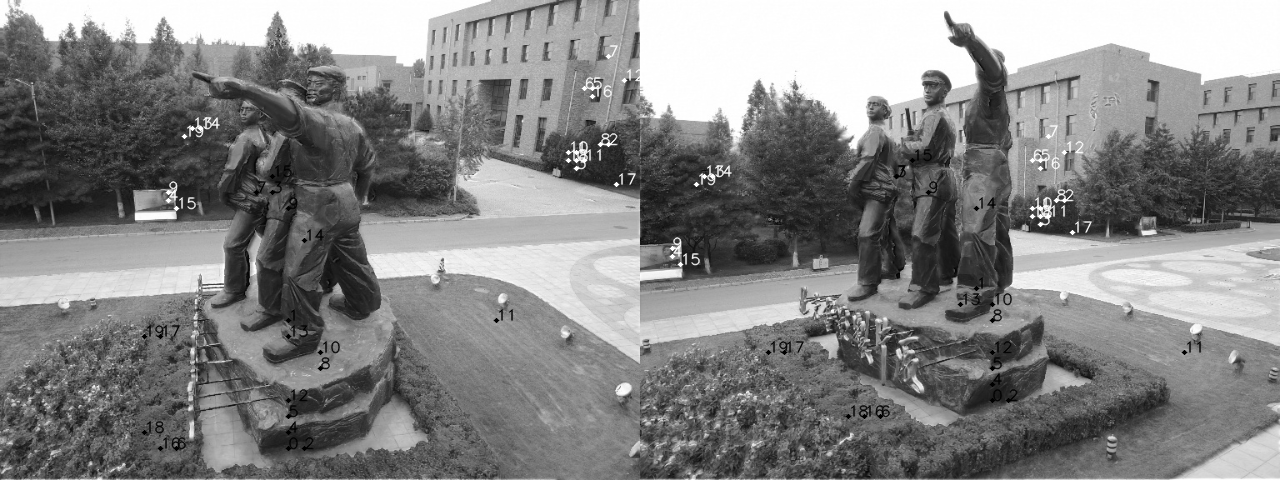}
	\caption{Dataset items processing results}
	\label{img:match1}
\end{figure}

The figure \ref{img:match1} shows examples of the model results on dataset images. The white dots indicate the matching results of the coarse LoFTR module of the original model, which was used as a teacher. And the black dots indicate the results of the smaller model. It is clear from the shown results that the smaller model focuses on different parts of images than the teacher model. The reason most likely is the smaller number of head layers, and different transformer parameters, that make the model emphasize more noticeable feature points. It is also possible to note the presence of errors in the presented feature matches for the smaller model, although in general feature matches are marked quite accurately.

\begin{figure}[ht]
	\includegraphics[width=\linewidth]{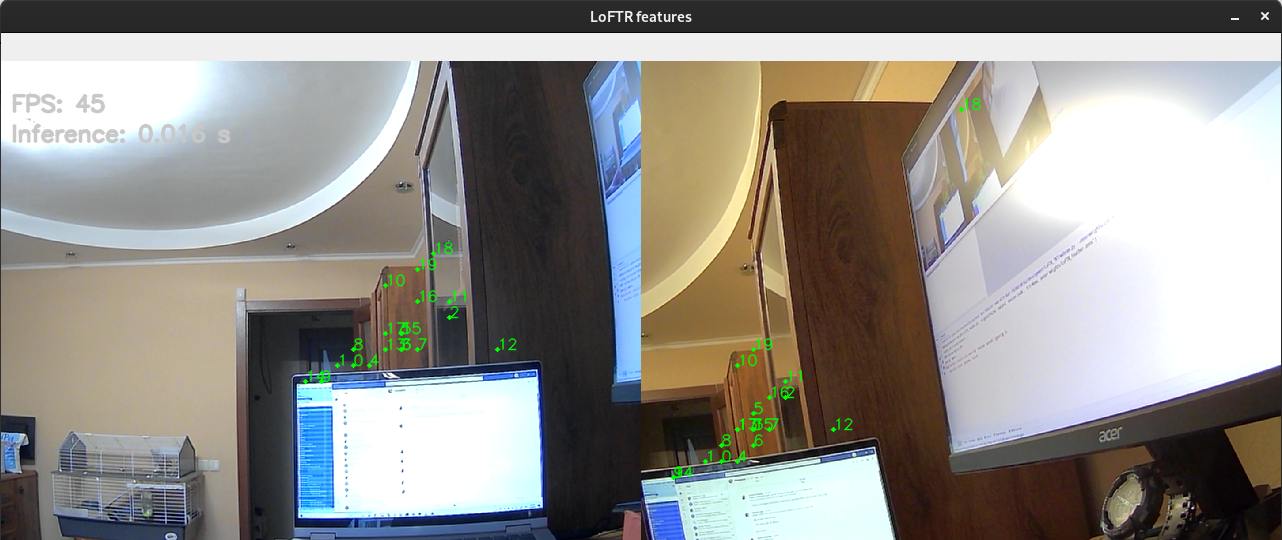}
	\includegraphics[width=\linewidth]{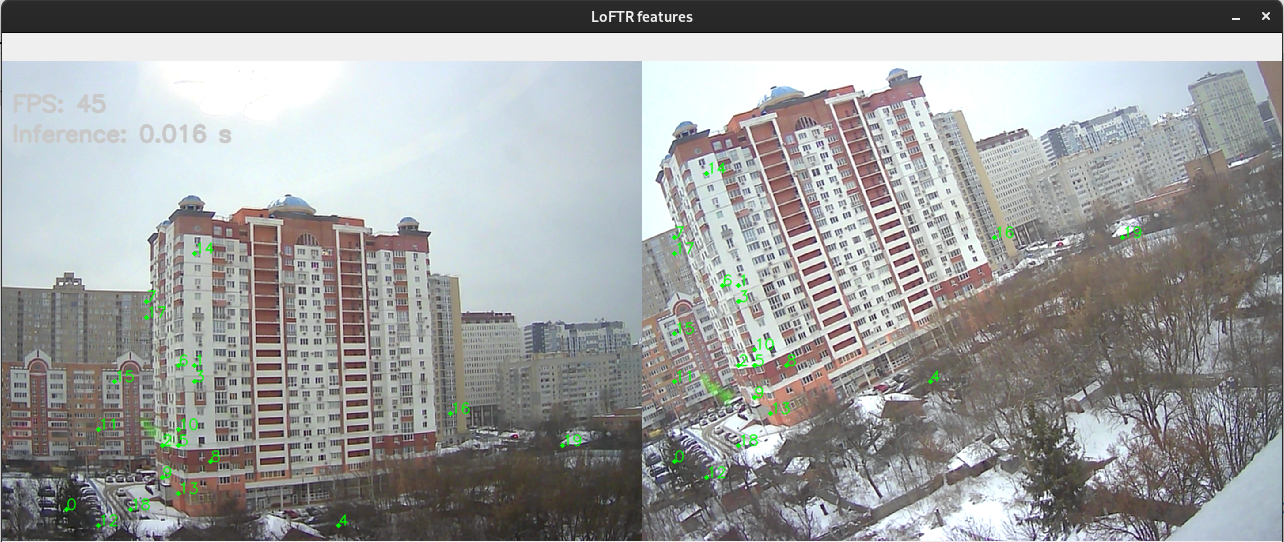}
	\includegraphics[width=\linewidth]{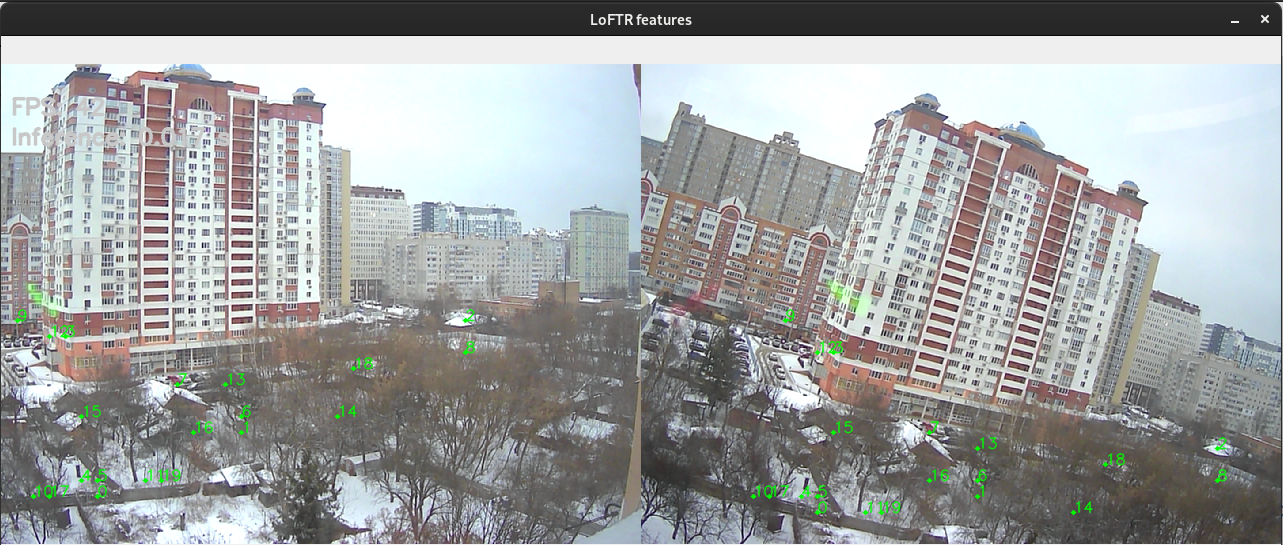}
	\caption{Realtime camera capture processing results}
	\label{img:match2}
\end{figure}

If we consider the model results on the images captured with a camera in real-time \ref{img:match2}, we can note that the model is able to process outdoor and indoor scenes images too. These samples also show that despite the fact that the model pays more attention to texture-intensive features, it can successfully find correct matches in images with repetitive texture patterns, i.e. it works well with the global image context.

\section{Conclusion}
This paper shows the process of transforming a relatively large LoFTR model into a lighter one that can be used on low-end devices with limited computational resources. To solve this problem, the fine matching transformer block was removed from the base model architecture, the number of head layers was significantly reduced, and the transformer in the coarse matching block was lightened. Also, due to the choice of NVIDIA TensorRT as a more computationally efficient runtime, the model source code was partially rewritten using more basic algorithms. This update increased the size of the code and most likely resulted in a performance loss, but such an effect was compensated by a smaller number of parameters and a different execution environment. Also, note that such a code adaptation can be a time-consuming task for a developer, so it's better to write the code with the target platforms in mind. The knowledge distillation technique was used to train the lightweight model and showed that the training speed and accuracy significantly increase for training with a teacher. Due to the use of hardware with a limited computational performance the Automatic Mixed Precision(AMP) and gradient accumulation techniques were applied to minimize memory consumption and speed up computation during the training process. Also, the BlendedMVS dataset was used for training. It has a smaller size than the one used to train the base model. But for ground-truth matches generation, a similar approach was used. This approach is based on depth maps and not on textural or geometric features. But such type of ground-truth feature generation sometimes can cause incorrect feature matches due to insufficient depth resolution. During training, it was noted that despite the use of such an optimization algorithm as Adam, the use of the learning rate scheduler significantly accelerates the training process. 
In general, the applied approach showed that the adapted and lightened model is able to detect feature matches in poorly textured areas of images but less effective than the base model. However, it successfully detects feature matches in images with repetitive patterns. And it shows good accuracy and performance on low-end devices with limited computational resources.

\bibliographystyle{abbrv}
\bibliography{refs}

\end{document}